\documentclass[runningheads]{llncs}
\usepackage{graphicx}
\usepackage{mathrsfs}
\usepackage{xcolor}
\usepackage{subcaption}
\usepackage{float}
\usepackage[11pt]{moresize}
\usepackage[english]{babel}
\usepackage[utf8]{inputenc}
\usepackage{csquotes}
\usepackage{makecell}
\begin{document}
\title{Dense Crowds Detection and Counting with a Lightweight Architecture\thanks{This work was supported by the Mexican National Council of Science and Technology CONACYT, and the FORDECyT project 296737 “Consorcio en Inteligencia Artificial”.}}

\author{Javier Antonio Gonzalez-Trejo\inst{1} \and
Diego A. Mercado-Ravell\inst{1}}
\authorrunning{J. A. Gonzalez-Trejo \& D. A. Mercado-Ravell }
\institute{Centro de Investigación en Matemáticas CIMAT, campus Zacatecas, 
Avenida Lasec, Andador Galileo Galilei, Manzana 3 Lote 7,
Parque Quantum, Zacatecas 98160
Mexico\\
\url{https://www.cimat.mx/}\\
\email{\{javier.gonzalez, diego.mercado\}@cimat.mx}}
\maketitle              %
\begin{abstract}
In the context of crowd counting, most of the works have focused on improving the accuracy without regard to the performance leading to algorithms that are not suitable for embedded applications. In this paper, we propose a lightweight convolutional neural network architecture to perform crowd detection and counting using fewer computer resources without a significant loss on count accuracy. The architecture was trained using the Bayes loss function to further improve its accuracy and then pruned to further reduce the computational resources used. The proposed architecture was tested over the USF-QNRF achieving a competitive Mean Average Error of 154.07 and a superior Mean Square Error of 241.77 while maintaining a competitive number of parameters of 0.067 Million. The obtained results suggest that the Bayes loss can be used with other architectures to further improve them and also the last convolutional layer provides no significant information and even encourage over-fitting at training.

\end{abstract}
\section{Introduction}
Crowd detection and counting is an interesting research area thanks to its wide range of applications in real-life scenarios from monitoring large groups of people, to even counting clustered objects of any category for example: biological cells \cite{NIPS2010_4043}, number of vehicles  on a traffic yam \cite{sam2017switching}, counting cattle \cite{XU2020105300}, analyzing large groups of objects in dense agglomeration \cite{NIPS2010_4043}.

These tasks have interesting challenges that have driven the creation of automated counting and detection methods, for example, the change of perspective from the camera and the crowd and heavy occlusion by the size of the crowd, environmental objects among others \cite{kang2018beyond}. Different approaches, from classical machine learning up to the more recent Deep Learning based from which the community have found more success at solving them  \cite{ma2019bayesian}.

The classical approach that has been taken in the past was to detect and  count each object individually \cite{ge09_marked} \cite{7807241}. In this approach, the first step is to detect each person and then proceed to perform the counting over the patches found. The main issue with this kind of methods is that despite the fact that they work relatively well on low density crowd scenes, they tend to fail at the moment severe occlusions occur, illumination changes or simple because the persons have a small representation in the image, since they require enough spatial information to perform the count. As a result of this, the required computational resources are considerably higher, since the counting task is a byproduct of classification and detection. 
    
More modern approaches involve hand-craft features \cite{chan08_privac} like SIFT (Scale Invariant Feature Transform) \cite{kuchhold18_scale_adapt_real_time_crowd} or HOG (Histogram of Oriented Gradients) analysis \cite{8599382}. However, such algorithms ditch the localization information on the head annotation, and cannot be utilized any further for crowd localization and tracking, resulting additionally in a poor performance in low density crowd scenes.
    
As the research results advances through the years, density maps have proven to be a better option \cite{kang2018beyond} \cite{Liu_2019_CVPR}. Density map provide the count from crowded scenes, but also spatial information that can be utilized on localization and tracking. One of the first works to utilize the modern concept of density map is in \cite{NIPS2010_4043}, where the authors first annotated images with single point annotations over the center of the object to count, then, since this points were not a meaningful target for learning, they introduced the idea of using Gaussian kernels to spread such points to cover the majority of the object to count and trained a linear model with the MESA (Maximum Excess over SubArrays) distance. Modern approaches ditched these classical machine learning methods for the most part, in favor of deep neural networks (DNN) thanks to the success of them in the classification task \cite{krizhevsky2012imagenet} while keeping the density maps as learning targets.

One of the most popular of the modern architectures that makes use of DNN, is the Multi-Column Convolutional Neural Network (MCNN) architecture \cite{zhang16_singl_image_crowd_count_multi}. On this work, the authors first propose a new way to generate the ground truth density map by changing the correlation matrix $\sigma$ of the Gaussian kernel depending on the number of annotations surrounding the annotation that is being softened. This makes the ground truth density maps to better represent the human heads since at higher crowd density, the head size will be smaller, thus the $\sigma$ shrink accordingly. Also, they introduced the MCNN architecture with 3 different columns of different kernel for big, medium and small crowd densities.

More recent works make use of a the well-known VGG architecture \cite{simonyan2014deep}, which has been found to have great transfer learning capabilities, in order to create increase the complexity to obtain better accuracy\cite{ma2019bayesian} \cite{Liu_2019_CVPR}. Other works also make use of multitask learning by adding the crowd density classification task to improve the crowd counting accuracy \cite{sindagi17_cnn_based} which also increase the complexity of the architectures by adding fully connected layers.

As shown, these methods generally focus on the quality of the density maps to further improve the counting problem results thus increasing the number of parameters to run the algorithm and, by result, increasing the inference time. 

Due to this and the recent success of reducing the complexity of architectures for classification like MobileNet \cite{howard2017mobilenets} derived in efforts to create Convolutional Neural Networks (CNN) architectures of lower complexity while maintaining a reasonable accuracy in both crowd detection and counting. This architectures are referred as lightweight CNN. The requirements for an architecture to be lightweight are not formally defined in the literature, thus we define that any architecture that has less than 1 million parameters. As such, the first architecture with high impact on the research community was the MCNN, even though is not mentioned by the authors \cite{zhang16_singl_image_crowd_count_multi}. Works that followed used ideas from state-of-the-art density map generators, and modified them to implement a less complex version. Such is the case of the Compact Convolutional Neural Network (CCNN) which used the tree multi column structure from \cite{zhang16_singl_image_crowd_count_multi} reducing them to only one layer per column \cite{shi2020realtime}. Furthermore, works in the line of lightweight density map generators were interested in Unmanned Aerial Vehicles (UAV) applications, from which there is no longer need of a data center to make the inference \cite{tzelepi19_discr_analy_regul_light_deep_cnn_model} \cite{Liu_2019_CVPR}. Such methods reduced the complexity of the algorithm and the density map output, such as it no longer provided the individuals location with in the image, but only the located crowd blobs. Works that wanted to provide single individual location as the heavyweight counter parts, utilized the concept of multiple columns as MCNN while reducing them at only one layer per column while adding an inference header that produced the density map \cite{shi2020realtime}.

All of this methods still try to improve their performance analyzing only the way they learn or the architecture itself, not taking into account that the hand crafted architecture can still introduce some over-fitting even at that a low number of parameters, or that the labels may not be correctly placed at the center of the person's heads \cite{ma2019bayesian}.
    
In this work we use the CCNN architecture to further improve it by training it using the Bayes Loss instead of the Euclidean distance which, by itself, reaches competitive results against the state-of-the-art. Then, we pruned the channels in the architecture, testing different configurations to find the lowest number of parameters and the best accuracy improvement if any, which is not commonly done in the literature for further reducing the complexity of the proposed architectures. 
Finally, we train one last time the the Pruned CCNN architecture for fine-tuning reaching state-of-the-art results. 
    
The main contributions of this paper are listed as follows:
\begin{itemize}
    \item Results for CCNN in the UCF-QNRF dataset. 
    \item Three architectures trained using the Bayes Loss; a modified MobileNetV2 with the fully connected layers removed that we call Bayes Loss (BL) MobileNetV2, a BL CCNN and a Pruned CCNN fine tuned with the Bayes Loss function. 
    \item Improvement of the state-of-the-art MSE metric with a 241.77 in the UCF-QNRF dataset by using the Bayes Loss for training and fine-tuning after pruning.
\end{itemize}
    
The paper is structured in the following sections. In Section \ref{sec:methodology } the proposed methodology is presented. Section \ref{sec:experiments} details the network structure, along with the dataset employed to train and validate the proposed architectures, including some training details. Section \ref{sec:results} we discuss the obtained results, showing the performance accomplished in both accuracy and computational resources. Finally, conclusions and future work are discussed at Section \ref{sec:conclusions}.
    
\section{Methodology}
\label{sec:methodology }
In this paper, we aim to reduce as much as possible the computational complexity on the lightweight CCNN, while keeping its accuracy, or even improving it when possible. To achieve this, we first train the CCNN using the Euclidean distance as loss function in the UCF-QNRF dataset \cite{idrees2018composition} to set a baseline, since the work that proposes th CCNN does not provide the metric in that specific dataset \cite{shi2020realtime}. After that, we compare it to three different CNNs, Bayes Loss (BL) MobileNet V2,  BL CCNN and Pruned BL CCNN, the last achieving state-of-the-art results.

For BL MobileNetV2, we modify the MobileNet V2 architecture, attach it to a regression header and then train it using the Bayes Loss in the UCF-QNRF dataset. 

For BL CCNN, we trained it without significant modification using the Bayes Loss in the same dataset, recording its results and then, we prune and re-trained the resulting CCNN for lowering the complexity of the model while achieving good accuracy.
In the following sections, we provide a detailed description of the CNNs, loss functions and pruning techniques used in this work.

\subsection{Compact Convolutional Neural Network}
\begin{table}[ht!]
    \centering
    \caption{Structure of the CCNN and Pruned CCNN. The multi column structure is shared in both architectures. The notation is as follows: operation - kernel size - padding - channels in - channels out. The layer 7 was the most pruned since it lost $80\%$ of its channels in the Pruned CCNN architecture.}
    \begin{tabular}{|c|c|c|}
    \hline
    \textbf{CCNN}     & \textbf{Prunned CCNN} & Layer Number\\
    \hline
    \multicolumn{2}{|c|}{\makecell{(conv2D-9x9-4x4-3-10, maxpool-2)\\
    (conv2D-7x7-3x3-3-14, maxpool-2)\\
    (conv2D-5x5-2x2-3-16, maxpool-2)}} & 0\\
    \hline
    conv2D-3x3-1x1-40-40 & conv2D-3x3-1x1-40-38 & 1\\
    conv2D-3x3-1x1-40-60 & conv2D-3x3-1x1-38-60 & 2\\
    maxpool-2 & maxpool-2 & 3\\
    conv2D-3x3-1x1-60-40 & conv2D-3x3-1x1-60-38 & 4\\
    maxpool-2 & maxpool-2 & 5\\
    conv2D-3x3-1x1-40-20 & conv2D-3x3-1x1-38-20 & 6\\
    conv2D-3x3-1x1-20-10 & conv2D-3x3-1x1-20-2 & 7\\
    conv2D-1x1-1x1-10-1 & conv2D-1x1-1x1-2-1 & 8\\
    \hline
    \end{tabular}

    \label{tab:ccnn}
\end{table}

The CCNN architecture improves over the concept first given in \cite{zhang16_singl_image_crowd_count_multi}, where  a multi column architecture is used in order to use different kernel sizes over the same image to extract features that take into account the density of the crowd \cite{shi2020realtime}. Same as MCNN, three columns for big, low and middle sized crowds are used. 
CCNN major contributions are the reduction of the parameters number through reducing the columns up to one layer each, and improving the count accuracy by creating a deeper regression layer to generate the density maps, as shown in table \ref{tab:ccnn}.
In order to train the baseline CCNN we up-sampled the output of the model to the size of the input image following the recommendations in \cite{gao2019c}. For CCNN trained with the Bayes loss (BL CCNN), we used bilinear up-scaling for the output to be half the input image size and applied an absolute value function as recommended in \cite{ma2019bayesian}.

\subsection{MobileNetV2}
\begin{table}[ht!]
    \centering
    \caption{Structure of BL MobileNetV2. We removed the fully connected module and the average pooling from the CNN, then attached a regression layer which produces the density maps. Each row describes a layer in the architecture repeated $n$ times. Each layer has the $c_o$ number of output channels, $c_i$ number of input channels, $s$ representing the stride and $t$ for the expansion rate for the bottleneck modules. All the 2D Convolutional operations (conv2D) use a kernel size of $(3, 3)$ with the exception of the last one, which uses a kernel of $(1, 1)$. We highlight the changes and additions made in the architecture.}
    \begin{tabular}{c|c|c|c|c|c}
    \hline
    Operation     & $t$ & $c_i$ & $c_o$ & $n$ & $s$\\
    \hline
    conv2D & - & 3 & 32 & 1 & 2\\
    bottleneck & 1 & 32 & 16 & 1 & 1\\
    bottleneck & 6 & 16 & 24 & 2 & 2\\
    bottleneck & 6 & 24 & 32 & 3 & 2\\
    bottleneck & 6 & 32 & 64 & 4 & 2\\
    bottleneck & 6 & 64 & 96 & 3 & 1\\
    bottleneck & 6 & 96 & 160 & 3 & \textbf{1}\\
    bottleneck & 6 & 160 & 320 & 1 & 1\\
    conv2D & - & 320 & 1280 & 1 & 1\\
    \hline
    \textbf{upsamplingBiliniar 2x} & - & 1280 & 1280 & - & -\\
    \textbf{conv2D} & - & 1280 & 640 & 1 & 1\\
    \textbf{conv2D} & - & 640 & 320 & 1 & 1\\
    \textbf{conv2D} & - & 320 & 160 & 1 & 1\\
    \textbf{conv2D} & - & 160 & 1 & 1 & 1\\

    \hline
    \end{tabular}

    \label{tab:mobilenetv2}
\end{table}
MobileNetV2 is a lightweight CNN that performs better in the classification task while being lighter than its predecessor, MobileNet \cite{sandler2018mobilenetv2}. It uses a special module called bottleneck residual block, which replaces fully convolutions with a factorized, low dimensional version  with almost near the same accuracy while reducing the number of parameters \cite{sandler2018mobilenetv2}.

The bottleneck residual block is composed of three parts, the first one is a point-wise convolution that expands the $k$ channels of the input by an expansion rate $t$. Then, a depth-wise convolution with kernel size of $(3, 3)$ and $kt$ channels is applied, where each kernel is applied to a single channel of the input. Finally, we project back to a low dimension space with a point-wise convolution with number of channels equal to the fully convolution that the bottleneck replaces.

We used a pre-trained model to replace the VGG19 feature extractor that the Bayes loss proposed \cite{ma2019bayesian} having in mind that MobileNetV2 performed better on the classification task than VGG19.

To use the architecture, we removed the fully connected module and the first average pooling layer, and attached a fully convolutional regression module, that takes the features extracted by the MobileNetV2 front-end and produces the density map at $\frac{1}{8}$ of the original size. At the output of the model, we perform an absolute value operation as recommended in \cite{ma2019bayesian}.

\subsection{Ground truth density maps generation and loss function}
As stated in \cite{ma2019bayesian}, the use of the Euclidean distance as loss function in the counting and detecting task is not desirable, since it does not take into account the labels that are not exactly in the center of the head of each person, while also does not take into account that each pixel is related to it neighbors to form a head. In order to solve these challenges. the Bayes loss was introduced, which instead of taking the ground truth density maps as learning targets, it takes them as likelihoods, while at the same time training and evaluating per label instead of pixel, thus taking into account their relationship.
The density maps are generated using Gaussian kernels over the head annotations to be used long with the Bayes loss defined next \cite{ma2019bayesian}:
\begin{equation}\label{eq:1}
    p(\mathrm{x}_m|y_m) = \mathcal{N}(\mathrm{x}_m;\mathrm{z}_n, \sigma^{2}\mathbf{I}_{2x2})
\end{equation}
\begin{equation}\label{eq:2}
    p(y_n|\mathrm{x}_m) = \frac{p(\mathrm{x}_m|y_n)}{\sum^N_{n=1}p(\mathrm{x}_m|y_n)}
\end{equation}

\begin{equation}\label{eq:3}
    E[c_n] = \sum^M_{m=1}p(y_n|\mathrm{x}_m) D^{est}(\mathrm{x})
\end{equation}

\begin{equation}\label{eq:4}
    \mathrm{z}_0^m = \mathrm{z}_n^{m} + d\frac{\mathrm{x}_m - \mathrm{z}^m_n}{||\mathrm{x}_m - \mathrm{z}^m_n  ||_2} 
\end{equation}

\begin{equation}\label{eq:5}
    p(\mathrm{x}_m|y_0) = \mathcal{N}(\mathrm{x}_m;\mathrm{z}_0^m, \sigma^{2}1_{2x2})
\end{equation}
\begin{equation}\label{eq:6}
    p(y_0|\mathrm{x}_m) = \frac{p(\mathrm{x}_m|y_n)}{\sum^N_{n=1}p(\mathrm{x}_m|y_n) + p(\mathrm{x}_m|y_0)}
\end{equation}

\begin{equation}\label{eq:7}
    E[c_0] = \sum^M_{m=1}p(y_0|\mathrm{x}_m) D^{est}(\mathrm{x}_m)
\end{equation}

\begin{equation}\label{eq:8}
    \mathcal{L}^{loss}=\sum_{n=1}^N\mathcal{F}(1-E[c_n]) + \mathcal{F}(0-E[c_0])
\end{equation}
where $\mathrm{x}_m$ represents a single pixel location, and $M$ is the total number of pixels. Furthermore, $z_n$ denotes the location of an annotation label $y_n$ and the total number of image annotations is $N$. Note that $y_0$ is a especial case of annotation which represent the background. $\mathcal{N}$ represents a 2D Gaussian distribution at $\mathrm{x}_m$ with mean $\mathrm{z}_n$ and a isotropic covariance $\sigma^{2}I_{2x2}$; acting as another hyper-parameter to optimize.
$E[c_n]$ is the expected value of the $c_n$, which is the total count over the annotation $y_n$. $D^{est}(\mathrm{x}_m)$ is the estimated count given by the trained architecture at the pixel $\mathrm{x}_m$. The same applies when $n=0$ for the background pixels.
To obtain the likelihood of $\mathrm{x}_m$ given $y_0$, $z_n^m$ is the nearest point of a pixel $\mathrm{x}_m$. Also, $z_0^m$ is a dummy background point which acts as a background label with respect of a pixel $\mathrm{x}_m$ of the image. Moreover, $d$ is an hyper-parameter that defines the ratio of separation between the $z^m_0$ from $z^m_n$, in other words, the expected human head size.
$\mathcal{L}$ is the loss function, in this case the Bayes loss.
$\mathcal{F}()$ is a distance function, in this case, we use the $L_1$ norm or Manhattan Distance . Additionally, $c_n$ is the total count over the annotation $y_n$ where the ground truth count for any annotation is 1. Finally, $c_0$ is the total count over the background where the ground truth count is 0. See \cite{ma2019bayesian} for further details about the Bayes loss function and its implementation on a heavyweight architecture.

\subsection{Pruning}
\begin{table}[ht!]
    \centering
    \caption{Pruned layers in the Pruned BL CCNN architecture. We use the $l_n$ norm to rank the channels on a given layer and remove the percentage of channels that ranked the lowest. Only three layers could be pruned without affecting the original accuracy performance of the BL CCNN.}
    \begin{tabular}{c|c|c}
    \hline
Layer number & Percentage of channels to be removed & $l_n$ norm used \\
    \hline
    $1$ & $5\%$ & $1$\\
    $4$ & $5\%$ & $1$\\
    $7$ & $80\%$ & $2$\\
    \hline
    \end{tabular}

    \label{tab:prunning}
\end{table}

The goal of pruning is to increment the sparsity of a CNN reducing the number of parameters \cite{li2016pruning}. Even though the CNN that we use in this paper is designed to have a small number of parameters, it is still likely to have some irrelevant channels that do not impact the performance significantly. 

To perform the pruning in the CCNN architecture, first we trained it using the Bayes Loss \cite{ma2019bayesian} to further improve its accuracy in order to have a large accuracy threshold to work with. Then, we evaluated the channel relevance on each individual layer using the $l_n$ norm. In Table \ref{tab:prunning} we show the number of channels pruned from each layer and the norm used to identify the least important. 

Pruning using $l_n$ norm works by ranking the channels and specifying a percentage of lower channels to be pruned, the idea that surged in \cite{li2016pruning} is that the more near zero the norm of a channel is, the less relevant to the final inference is. This is no always truth, thus the need to try each channel separately and in groups since increasing the sparsity, and more important in lightweight models, could decrease the accuracy so far below making the architecture unviable. In our tests all but the layer 7 of the regression header had a negative impact on the performance of the CCNN architecture, having the layer 2 with the greater impact. The layer 7 had almost the same accuracy in the validation set, even after removing up to 6 channels. At 8 channels removed from layer 7, we can notice an increase on the architecture performance, which might be due to this channel introducing over-fitting.
Once we obtained a slightly better result on the validation set, we trained the Pruned CCNN again for fine-tuning, which yielded better results than the BL CCNN, both in accuracy and in computational complexity, as shown in the Table \ref{tab:accuracy}.

\section{Experiments}
\label{sec:experiments}

\subsection{Evaluation metrics}
We used the two most widely used metrics for crowd counting: Mean Absolute Error (MAE) and
Mean Squared Error (MSE) which are defined as follows \cite{kang2018beyond}:
\begin{equation}
    MAE =  \frac{1}{K} \sum^K_{k=1} |N_k - \hat{N}_k|
\end{equation}
\begin{equation}
    MSE = \sqrt{\frac{1}{k}\sum^K_{k=1} |N_k - \hat{N}_k|^2}
\end{equation}
where $N_k$ represents the total number of persons on an image $k$, with $K$ number of images on the training set, and $\hat{N}_k $ is the estimated number of persons inferred from the model.\\
These metrics are needed to analyze for how much the count differs from the ground truth and the variability of the errors. MAE, since is not sensitive to outliers, is used to compare the global error between architectures, while MSE is used to compare which method is more prone to make big mistakes, since it gives more weight to big differences between ground truth count and the estimated count.

\subsection{UCF-QNRF dataset}
The public dataset comprehends a set of $1535$ images spliced in $1201$ for training and $334$ for testing which represents roughly $80\%$ and $20\%$ respectively. The annotations are single points in the center of the visible part of the head of each person in the image. The average count is $815$ persons per image, having $12865$ persons per image as the upper limit. The dataset has more challenging scenarios than traditional crowd datasets like ShanghaiTech \cite{zhang16_singl_image_crowd_count_multi}  and WorldExpo'10 \cite{zhang2015cross} since it has images obtained from the web including sites as Flickr, thus having considerably more variety with regards to locations, backgrounds and illumination, among others. More importantly, the images resolution for the majority of the dataset is higher than $2101 \times 2888$ pixels, which helped the human experts to label with more precision each head \cite{idrees2018composition}.

\subsection{Training details}
Next, we describe the hyper parameters that we used to train each of the architectures here presented. For all of them, we fine tuned the hyperparameters using a validation set of $100$ images, randomly obtained from the train set of the UCF-QNRF dataset.
    
The optimizer used for all the architectures was the Adam algorithm seeing that is the most used for both classification and crowd counting task \cite{kingma2014adam}. The hyper parameters concerning the Bayes loss in the likelihood function in Eq.(\ref{eq:1}) $\sigma$, and in the $z_0^m$ function generator in Eq. (\ref{eq:4}) $d$, where set to 12.0 and 0.1, respectively. Since the images from the dataset are from different scenes, we set the batch size to 1 in order to not introduce bias and variance from mini batches that do not have images of the same scene, thus leading to not generalizing over unseen data and not converging fast enough.

For the baseline CCNN, we set the learning rate to 0.0001, weight decay to 0.00004 and  $\sigma = 15$ to generate the ground truth density maps.

For BL MobileNetV2, we use horizontal flipping for data augmentation, at the $100th$ epoch we started to validate the model every 5 epochs saving the best model, learning rate set to 0.0001 and weight decay set to 0.00004. The architecture was filled with pre-trained weights in the ImageNet dataset in order to get low level features. The regression header was initialized using the Kaiming normal \cite{he2015delving}.

For BL CCNN, we use random crop of 512 by 512 pixels and horizontal flapping and we started to validate our results at the $100th$ epoch at each $5$ epocj saving the best model . We use Adam optimizer with learning rate set to $0.0001$, and weight decay set to $0.0001$. Hyperparameters concerning the Bayes loss, like the standard deviation $\sigma$ and the background ratio $d$, where set to $8.0$ and $15\%$ respectively. Testing with values of $12$ and $10\%$ for $\sigma$ and background ratio $d$ were done since according \cite{ma2019bayesian} these values could lead to better accuracy, but we found that they lead the model to not converging at reasonable accuracy in the validation test. We used the same hyperparameters for Pruned BL CCNN in the fine-tuning step.

\section{Results and discussion}
\label{sec:results}
\begin{table}[ht!]    

\caption{ Benchmark comparison of state-of-the-art lightweight architectures for crowd counting, using the dataset UCF-QNRF. The MAE and MSE metrics are used for accuracy comparison, and number of parameters for complexity comparison. From the studied architectures, the Pruned BL CCNN proposed in this work has the best MSE of 241.77, a competitive MAE of 154.07 while providing a competitive number of parameters of $0.067$ M.}
\label{tab:accuracy}
\centering
\begin{tabular}{|l|l|l|l|}
\hline
Name & MAE & MSE & Parameters\\
\hline
MCNN \cite{zhang16_singl_image_crowd_count_multi} & $315.00$ &  $508.00$ & $0.13$ M\\
CCNN \cite{shi2020realtime}& $224.20$ & $331.00$ & $0.072$ M\\
RHNet \cite{yu19_rhnet}& $172.00$ & $258.00$ & \textbf{0.03 M}\\
1/4-SANet + SKT \cite{liu2020efficient}& $157.46$ & $257.66$ & $0.058$ M\\
PCC Net \cite{gao2019pcc} & $148.70$ & $247.30$ & $0.55$ M \\
SANet \cite{cao2018scale} & \textbf{152.59} & $246.98$ & $0.91$ M\\
\textbf{BL MobileNetV2 (Ours)}& $230.11$ & $388.01$ & $11.90$ M \\
\textbf{BL CCNN (Ours)} & $172.67$ &  $272.55$ & $0.072$ M\\
\textbf{Pruned BL CCNN (Ours)} & $154.07$ & \textbf{241.77} & $0.067$ M\\
\hline
\end{tabular}
\end{table}

\begin{figure}[ht!]
    \centering
    \begin{tabular}{cccc}
        \subcaptionbox{GT Count: 350\label{a}}{\includegraphics[width = 1.16in]{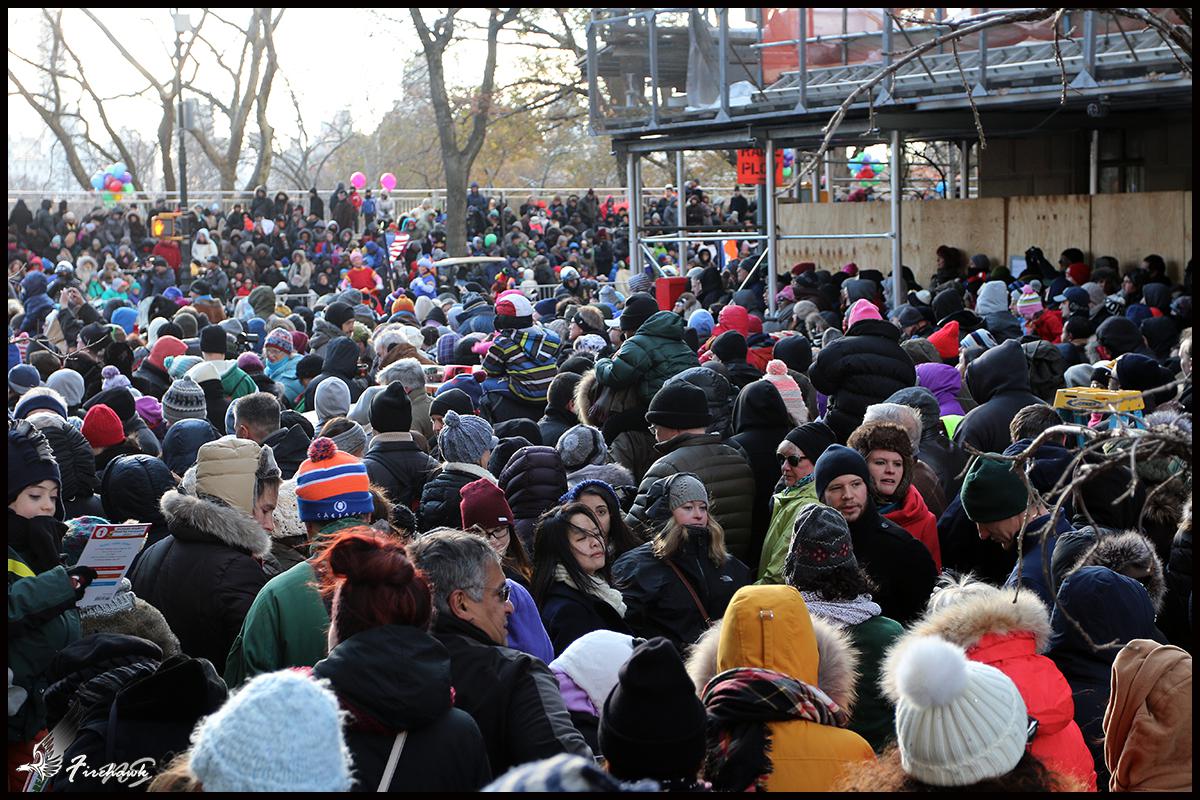}} &
        \subcaptionbox{Inferred: 335.65\label{c}}{\includegraphics[width = 1.16in]{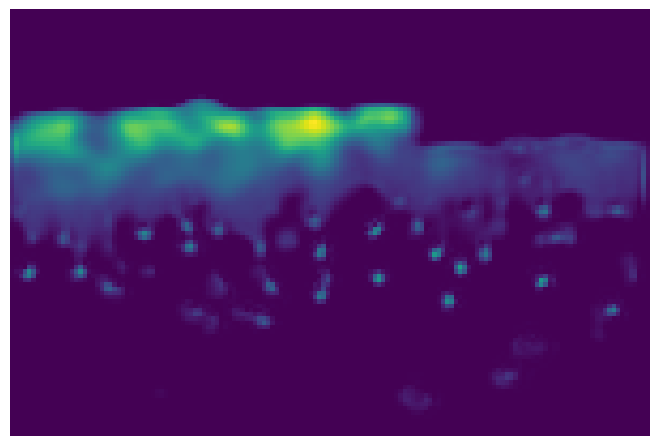}} &
        \subcaptionbox{Inferred: 317.23\label{b}}{\includegraphics[width = 1.16in]{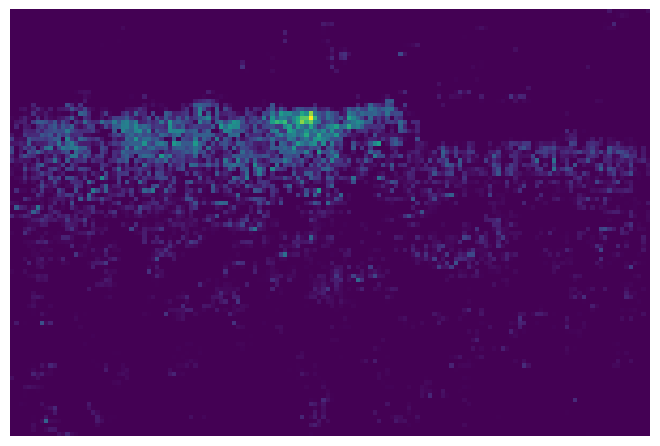}} &
        \subcaptionbox{Inferred: 354.86\label{d}}{\includegraphics[width = 1.16in]{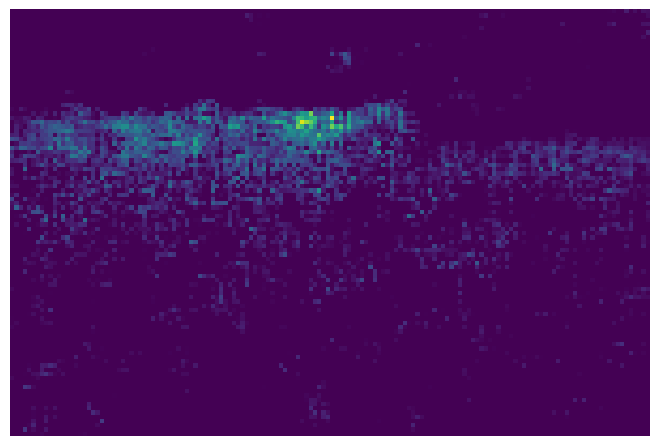}}\\
        
        \subcaptionbox{GT Count: 434\label{e}}{\includegraphics[width = 1.16in]{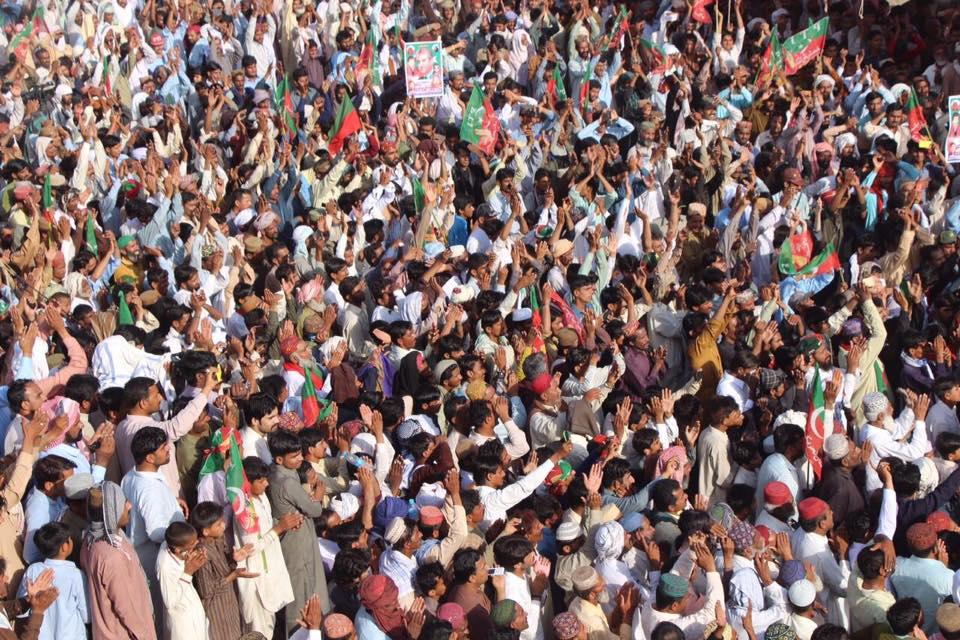}} &
        \subcaptionbox{Inferred: 167.80\label{f}}{\includegraphics[width = 1.16in]{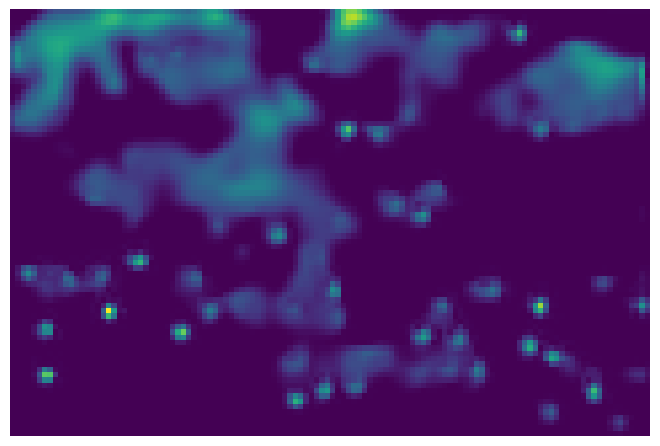}} &
        \subcaptionbox{Inferred: 381.60\label{g}}{\includegraphics[width = 1.16in]{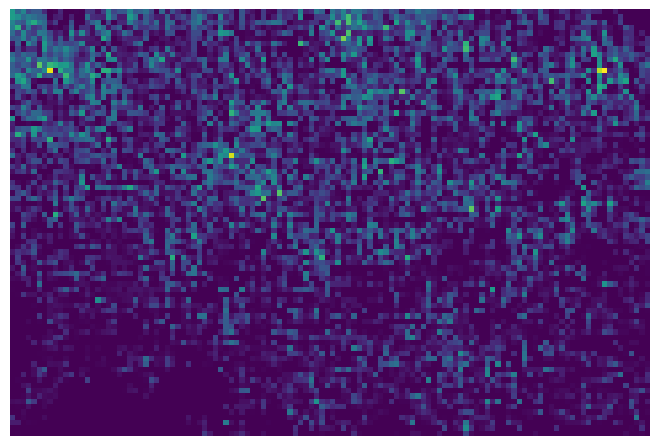}} &
        \subcaptionbox{Inferred: 443.06\label{h}}{\includegraphics[width = 1.16in]{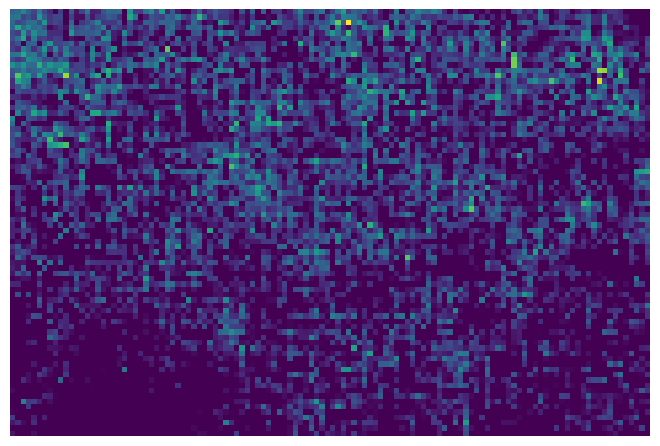}}\\
        
        \subcaptionbox{GT Count: 225\label{i}}{\includegraphics[width = 1.16in]{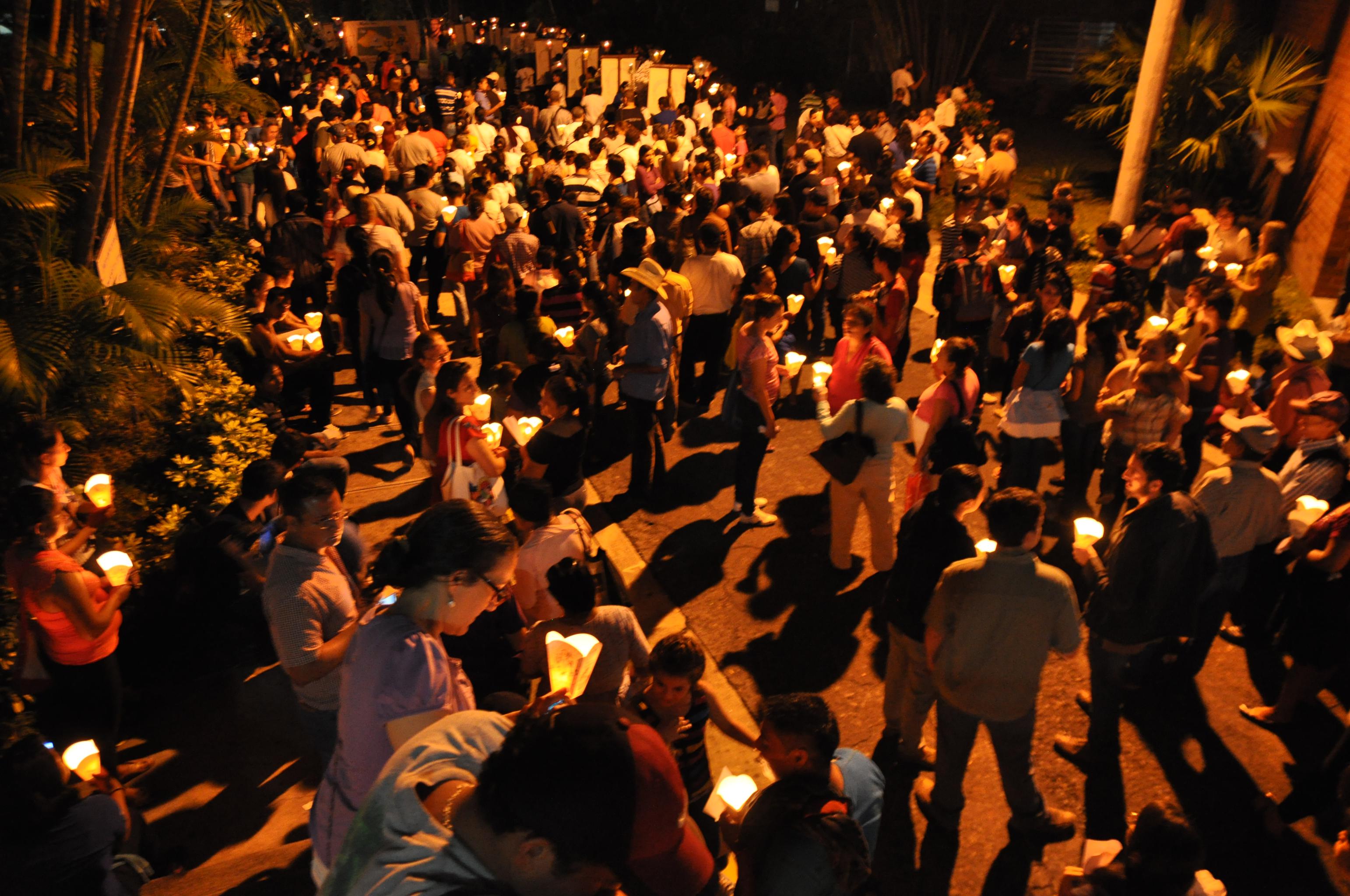}} &
        \subcaptionbox{Inferred: 420.94\label{j}}{\includegraphics[width = 1.16in]{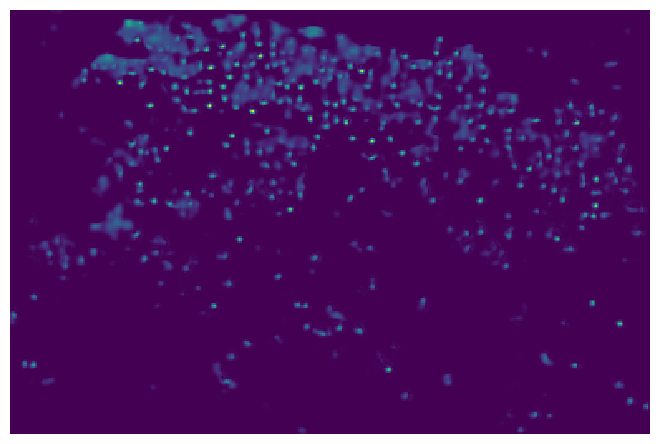}} &
        \subcaptionbox{Inferred: 319.29\label{k}}{\includegraphics[width = 1.16in]{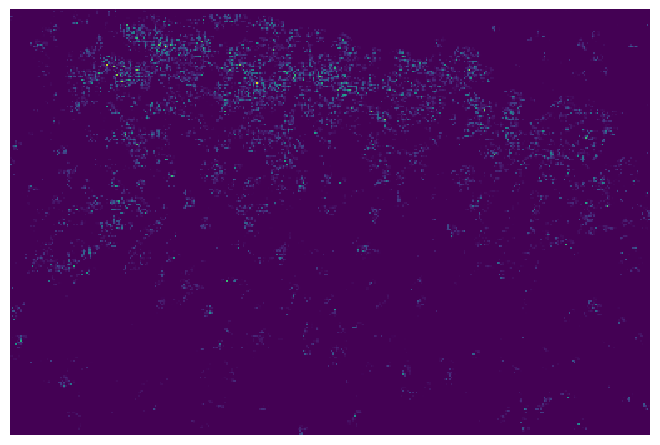}} &
        \subcaptionbox{Inferred: 226.43\label{l}}{\includegraphics[width = 1.16in]{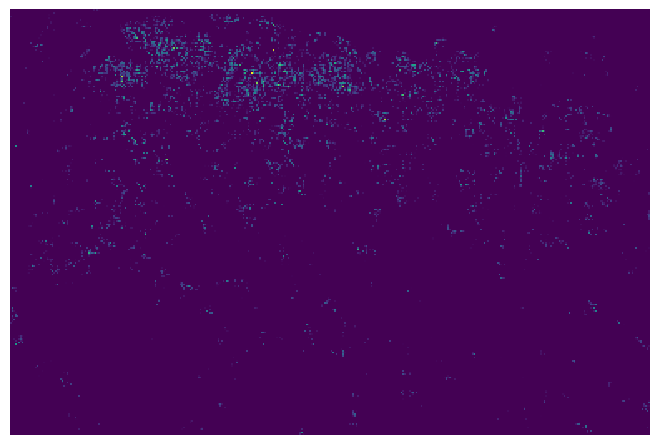}}\\
        
        \subcaptionbox{GT Count: 1018\label{m}}{\includegraphics[width = 1.16in]{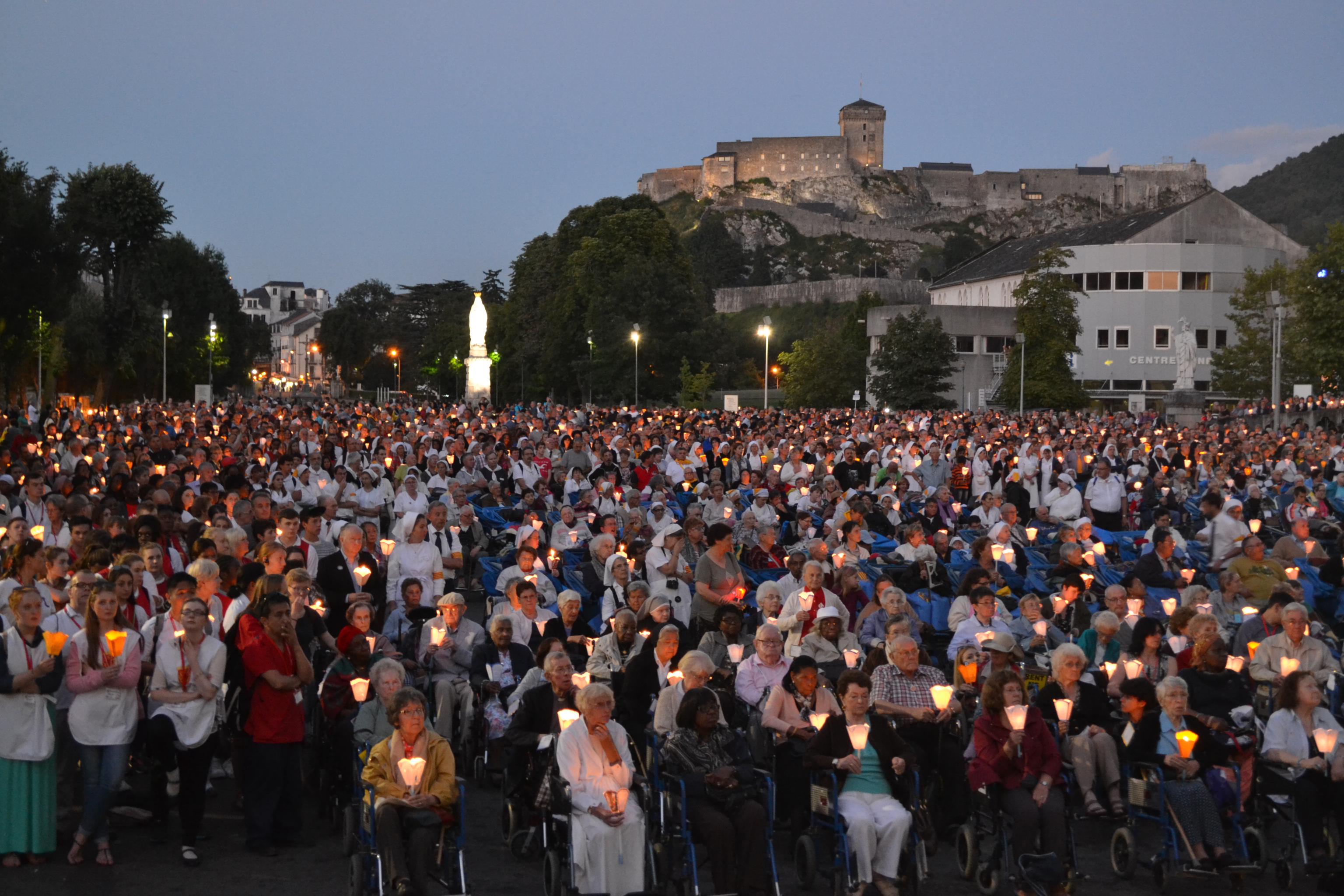}} &
        \subcaptionbox{Inferred: 1056.55\label{n}}{\includegraphics[width = 1.16in]{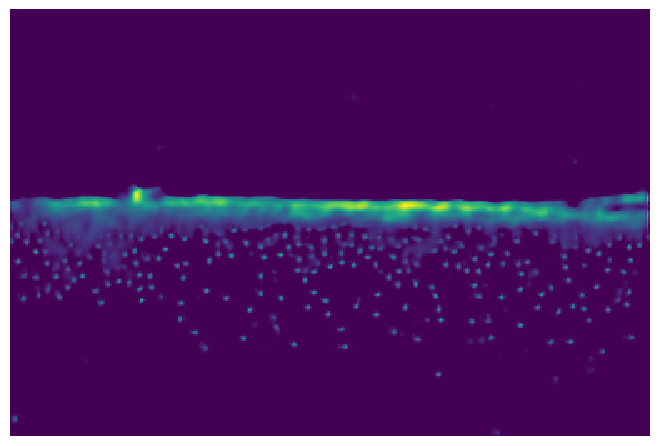}} &
        \subcaptionbox{Inferred: 826.27\label{o}}{\includegraphics[width = 1.16in]{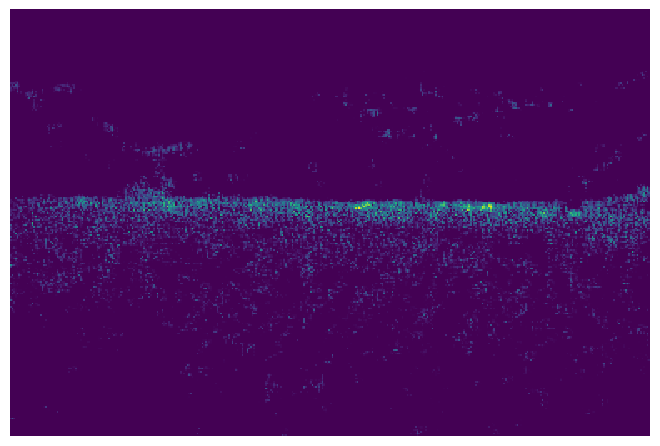}} &
        \subcaptionbox{Inferred: 945.00\label{p}}{\includegraphics[width = 1.16in]{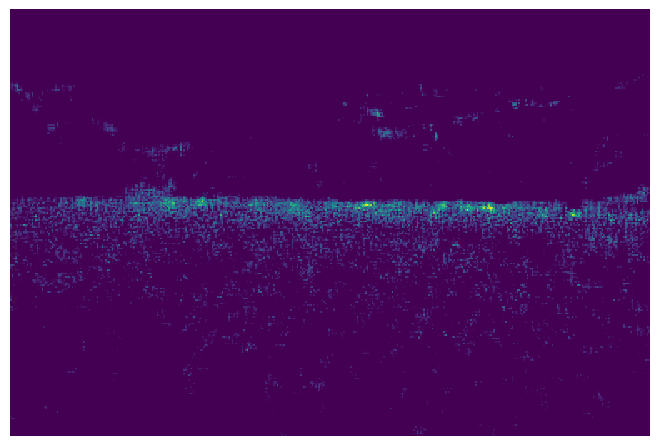}}\\
        
        \subcaptionbox{GT Count: 201\label{q}}{\includegraphics[width = 1.16in]{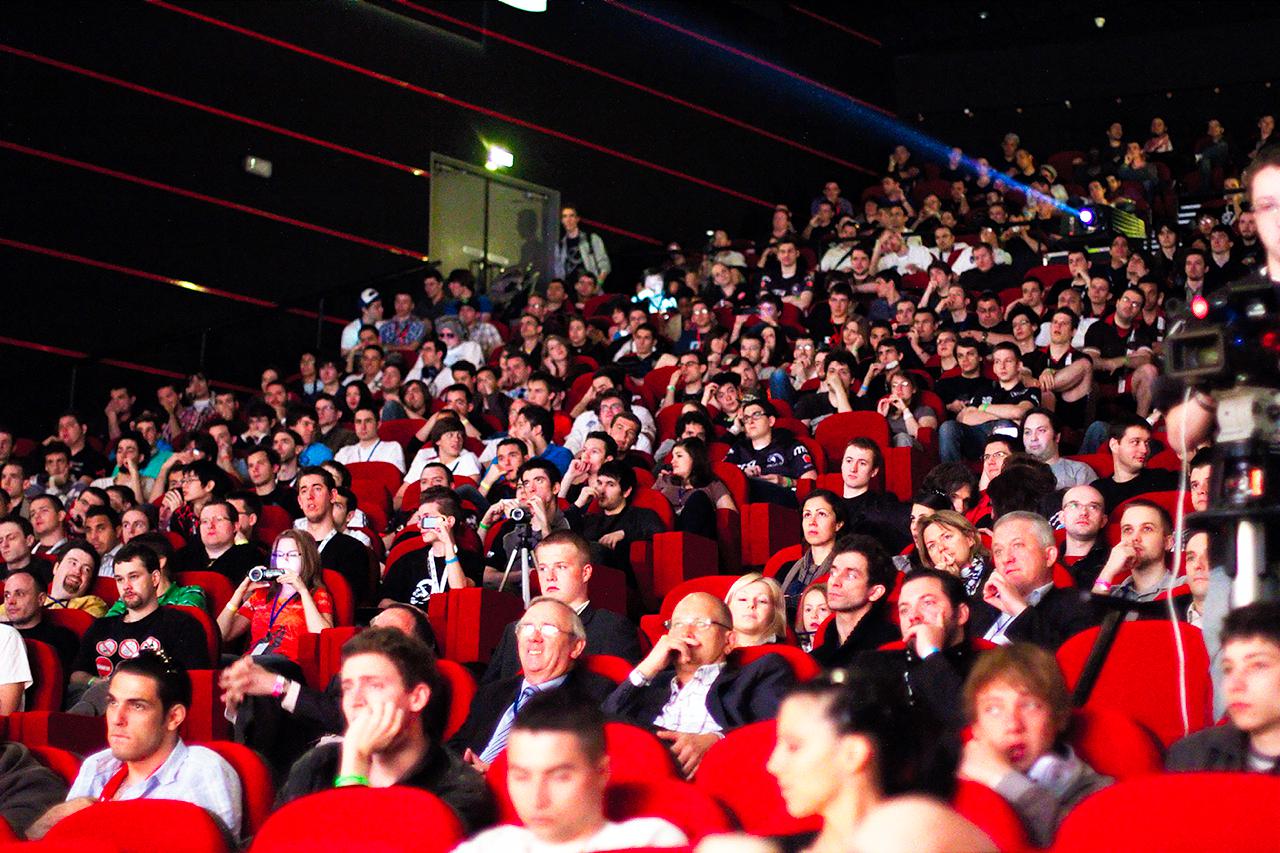}} &
        \subcaptionbox{Inferred: 209.64\label{r}}{\includegraphics[width = 1.16in]{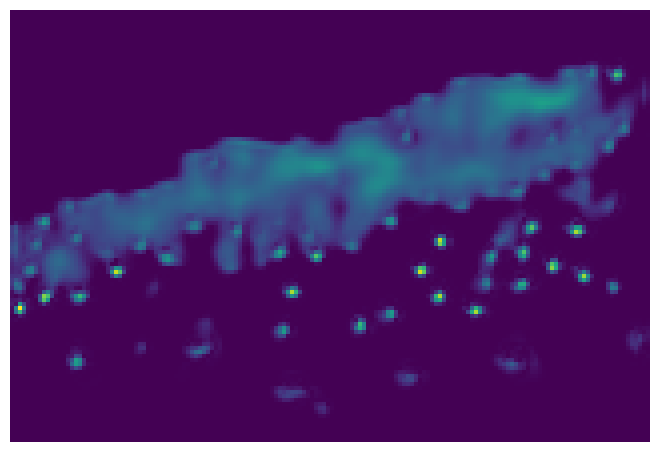}} &
        \subcaptionbox{Inferred: 189.48\label{s}}{\includegraphics[width = 1.16in]{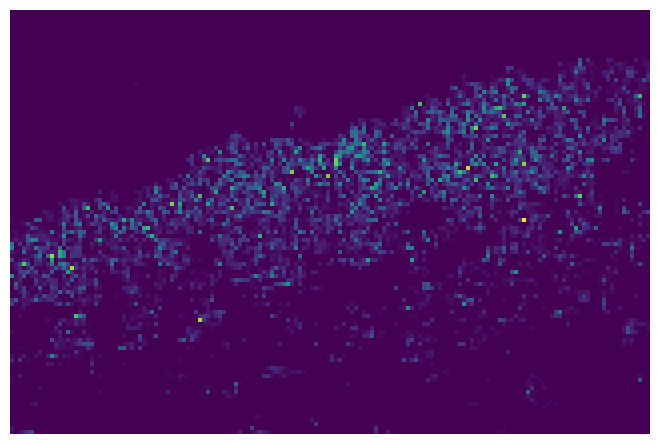}} &
        \subcaptionbox{Inferred: 206.11\label{t}}{\includegraphics[width = 1.16in]{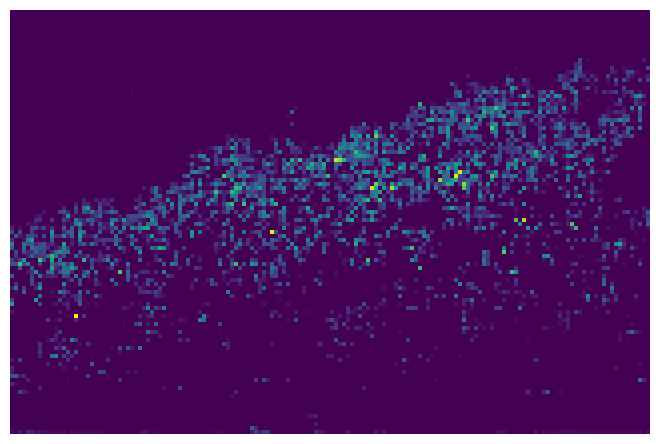}}\\
        \ssmall{(u)Images Samples} &
        \ssmall{(v)BL MobileNet V2} &
        \ssmall{(w)BL CCNN} &
        \ssmall{(x)Pruned BL CCNN}\\
    \end{tabular}
    \caption{Density maps generated by (v) BL MobileNetV2, (w) BL CCNN and (x) Pruned BL CCNN. All of the images are 1/8 of the size of the original image input. The more yellowish the higher person count. The better individual localization results come from BL MobileNetV2, and the better count comes from Pruned BL CCNN.}
    \label{fig:density_comparition}
\end{figure}

We compare the obtained results against the state-of-the-art on lightweight density map generators. In Table \ref{tab:accuracy} we show the MAE, MSE and Parameters number of some of these architectures and compare them against our proposed algorithms. Also, in Figure \ref{fig:density_comparition} we provide a visual comparison of the quality of density map generated by our three density map generators in images obtained from the test set of the UCF-QNRF dataset.

For a fair comparison, and since in \cite{shi2020realtime} they did not use the UCF-QNRF dataset, we trained the CCNN architecture using the Euclidean distance as loss function and Adam optimizer with learning rate of $0.0005$ which yielded results only slightly better than MCNN.

For our first architecture, we used the same idea of implementing a pretrained architecture like VGG with a regressor header to be trained with the Bayes Loss function, but using a lightweight architecture, in this case MobileNetV2. As we can see in the Table \ref{tab:accuracy}, BL MobileNetV2 is only better than MCNN while having at least 10 times more the parameters which lead us to believe that transfer learning for lightweight architectures for the classification task are not as straight forward as VGG. Furthermore, in the Figure \ref{fig:density_comparition} in the column of BL MobileNetV2, we can see that compared whit our other architectures for individuals near the camera, the localization performs better than the others. Specifically, in Figure \ref{r} we can clearly see where the individuals of the first row of the image are located compared with Figures \ref{s} and \ref{t} where is not clearly identified where an individual is located and where a crowd starts.

For our second architecture, BL CCNN, we obtained 172.67 for MAE and 272.55 for MSE, much better compared with CCNN by only using the Bayes Loss function. Before reaching that results, we tested different values of the Bayes Loss hyperparameters. We tested the combinations for $\sigma$ ($8$ and $12$) and $d$ ($0.10$ and $0.15$). The combination of $\sigma = 8$ and $d = 0.15$ proved to be the only one to provide the results provided while any other combination yielded $500$ for MAE or above in the validation test, which, at least for the CCNN architecture, invalidates what is provided in \cite{ma2019bayesian}. In regard of the quality of the density maps, we found that the model does not provide information about individual localization. In the Figures \ref{c}, \ref{g}, \ref{k}, \ref{o} and \ref{s} we see that each provides where the crowd density concentrates in the image while being the architecture that underestimates the most.

Before our third architecture, we found that pruning only the layer 7 of Table \ref{tab:prunning} was enough to increase the accuracy of the CCNN architecture. In the validation test CCNN got 185.4 for MAE 306.10 for MSE, pruning the layer 7 yielded 177.26 for MAE and 300.8 for MSE and finally the fully Pruned CCNN yielded 187.019 MAE and 307 for MSE. Pruning any other layer yield worst results having the layer $2$ the most important for the inference.

Finally, our Pruned BL CCNN obtained state-of-the-art MSE results of 241.77 while obtaining competitive results of 154.07 of MAE and 0.067 Million of parameters. These results indicate that our architecture yield less big errors compared with the other architectures, while being the third most lightweight of them. In Figure \ref{fig:density_comparition} in the column for Pruned BL CCNN, we can see that only the counting performed better. This is in line with the work of \cite{kang2018beyond} that tell us that a better individual localization yields worst counting accuracy and vice versa. 

\section{Conclusions and Future Work}
\label{sec:conclusions}
In this paper, we present three different architectures, a modified MobileNetV2 with a regressed header trained with Bayes Loss, the CCNN trained with Bayes Loss and a Pruned version of CCNN fine tuned with the Bayes Loss which provided a state-of-the-art MSE of 241.77 while providing the third lowest number of parameters of the compared architectures.  
We conclude that while VGG19 can be used for transfer learning for the crowd counting task, further work has to be done for using lightweight models like MobileNetV2 in order to offer comparable accuracy results while having a smaller number of results.

For CCNN, the Bayes Loss function provides a substantial improvement of counting accuracy only for the values $\sigma=8$ and $d = 0.15$ with accuracy of 172.67 for MAE and 272.55 for MSE compared with the poor results of CCNN trained with the Euclidean loss of 224.20 for MAE and 331.00 for MSE.

And for the Pruned BL CCNN, we found that even though the CCNN is designed to be lightweight, the layer second to last can be pruned by an $80\%$ while providing better results than the original. Finally, that the combination of pruning with a superior Bayes Loss function for fine-tuning is a good combination for increasing the accuracy and decreasing the number of parameters of lightweight architectures.

In future work, we aim to implement the architecture embedded in a drone and test it in real world scenarios. Finally, we will like to create a custom lightweight CNN that provides better individual localization while still providing state-of-the-art counting accuracy.

\bibliographystyle{splncs04}
\bibliography{references}

\begin{thebibliography}{10}
\providecommand{\url}[1]{\texttt{#1}}
\providecommand{\urlprefix}{URL }
\providecommand{\doi}[1]{https://doi.org/#1}

\bibitem{cao2018scale}
Cao, X., Wang, Z., Zhao, Y., Su, F.: Scale aggregation network for accurate and
  efficient crowd counting. In: Proceedings of the European Conference on
  Computer Vision (ECCV). pp. 734--750 (2018)

\bibitem{chan08_privac}
Chan, A.B., Liang, Z.S.J., Vasconcelos, N.: Privacy preserving crowd
  monitoring: Counting people without people models or tracking. In: 2008 IEEE
  Conference on Computer Vision and Pattern Recognition (6 2008).
  \doi{10.1109/cvpr.2008.4587569},
  \url{https://doi.org/10.1109/cvpr.2008.4587569}

\bibitem{gao2019c}
Gao, J., Lin, W., Zhao, B., Wang, D., Gao, C., Wen, J.: C$^3$ framework: An
  open-source pytorch code for crowd counting. arXiv preprint arXiv:1907.02724
  (2019)

\bibitem{gao2019pcc}
Gao, J., Wang, Q., Li, X.: Pcc net: Perspective crowd counting via spatial
  convolutional network (2019)

\bibitem{ge09_marked}
Ge, W., Collins, R.T.: Marked point processes for crowd counting. In: 2009 IEEE
  Conference on Computer Vision and Pattern Recognition. p.~nil (6 2009).
  \doi{10.1109/cvpr.2009.5206621},
  \url{https://doi.org/10.1109/cvpr.2009.5206621}

\bibitem{he2015delving}
He, K., Zhang, X., Ren, S., Sun, J.: Delving deep into rectifiers: Surpassing
  human-level performance on imagenet classification (2015)

\bibitem{howard2017mobilenets}
Howard, A.G., Zhu, M., Chen, B., Kalenichenko, D., Wang, W., Weyand, T.,
  Andreetto, M., Adam, H.: Mobilenets: Efficient convolutional neural networks
  for mobile vision applications. arXiv preprint arXiv:1704.04861  (2017)

\bibitem{idrees2018composition}
Idrees, H., Tayyab, M., Athrey, K., Zhang, D., Al-Maadeed, S., Rajpoot, N.,
  Shah, M.: Composition loss for counting, density map estimation and
  localization in dense crowds. In: Proceedings of the European Conference on
  Computer Vision (ECCV). pp. 532--546 (2018)

\bibitem{kang2018beyond}
Kang, D., Ma, Z., Chan, A.B.: Beyond counting: Comparisons of density maps for
  crowd analysis tasks—counting, detection, and tracking. IEEE Transactions
  on Circuits and Systems for Video Technology  \textbf{29}(5),  1408--1422
  (2018)

\bibitem{kingma2014adam}
Kingma, D.P., Ba, J.: Adam: A method for stochastic optimization (2014)

\bibitem{krizhevsky2012imagenet}
Krizhevsky, A., Sutskever, I., Hinton, G.E.: Imagenet classification with deep
  convolutional neural networks. In: Advances in neural information processing
  systems. pp. 1097--1105 (2012)

\bibitem{kuchhold18_scale_adapt_real_time_crowd}
Kuchhold, M., Simon, M., Eiselein, V., Sikora, T.: Scale-adaptive real-time
  crowd detection and counting for drone images. In: 2018 25th IEEE
  International Conference on Image Processing (ICIP). p.~nil (10 2018).
  \doi{10.1109/icip.2018.8451289},
  \url{https://doi.org/10.1109/icip.2018.8451289}

\bibitem{NIPS2010_4043}
Lempitsky, V., Zisserman, A.: Learning to count objects in images. In:
  Lafferty, J.D., Williams, C.K.I., Shawe-Taylor, J., Zemel, R.S., Culotta, A.
  (eds.) Advances in Neural Information Processing Systems 23, pp. 1324--1332.
  Curran Associates, Inc. (2010),
  \url{http://papers.nips.cc/paper/4043-learning-to-count-objects-in-images.pdf}

\bibitem{li2016pruning}
Li, H., Kadav, A., Durdanovic, I., Samet, H., Graf, H.P.: Pruning filters for
  efficient convnets (2016)

\bibitem{7807241}
{Liu}, C., {Chen}, H., {Lo}, K., {Wang}, C., {Chuang}, J.: Accelerating
  vanishing point-based line sampling scheme for real-time people localization.
  IEEE Transactions on Circuits and Systems for Video Technology
  \textbf{27}(3),  409--420 (2017)

\bibitem{liu2020efficient}
Liu, L., Chen, J., Wu, H., Chen, T., Li, G., Lin, L.: Efficient crowd counting
  via structured knowledge transfer (2020)

\bibitem{Liu_2019_CVPR}
Liu, W., Salzmann, M., Fua, P.: Context-aware crowd counting. In: The IEEE
  Conference on Computer Vision and Pattern Recognition (CVPR) (June 2019)

\bibitem{ma2019bayesian}
Ma, Z., Wei, X., Hong, X., Gong, Y.: Bayesian loss for crowd count estimation
  with point supervision. In: Proceedings of the IEEE International Conference
  on Computer Vision. pp. 6142--6151 (2019)

\bibitem{sam2017switching}
Sam, D.B., Surya, S., Babu, R.V.: Switching convolutional neural network for
  crowd counting. In: 2017 IEEE Conference on Computer Vision and Pattern
  Recognition (CVPR). pp. 4031--4039. IEEE (2017)

\bibitem{sandler2018mobilenetv2}
Sandler, M., Howard, A., Zhu, M., Zhmoginov, A., Chen, L.C.: Mobilenetv2:
  Inverted residuals and linear bottlenecks. In: Proceedings of the IEEE
  conference on computer vision and pattern recognition. pp. 4510--4520 (2018)

\bibitem{shi2020realtime}
{Shi}, X., {Li}, X., {Wu}, C., {Kong}, S., {Yang}, J., {He}, L.: A real-time
  deep network for crowd counting. In: ICASSP 2020 - 2020 IEEE International
  Conference on Acoustics, Speech and Signal Processing (ICASSP). pp.
  2328--2332 (2020)

\bibitem{simonyan2014deep}
Simonyan, K., Zisserman, A.: Very deep convolutional networks for large-scale
  image recognition (2014)

\bibitem{sindagi17_cnn_based}
Sindagi, V.A., Patel, V.M.: Cnn-based cascaded multi-task learning of
  high-level prior and density estimation for crowd counting. In: 2017 14th
  IEEE International Conference on Advanced Video and Signal Based Surveillance
  (AVSS) (8 2017). \doi{10.1109/avss.2017.8078491},
  \url{https://doi.org/10.1109/avss.2017.8078491}

\bibitem{tzelepi19_discr_analy_regul_light_deep_cnn_model}
Tzelepi, M., Tefas, A.: Discriminant analysis regularization in lightweight
  deep cnn models. In: 2019 IEEE International Conference on Image Processing
  (ICIP) (9 2019). \doi{10.1109/icip.2019.8803676},
  \url{https://doi.org/10.1109/icip.2019.8803676}

\bibitem{XU2020105300}
Xu, B., Wang, W., Falzon, G., Kwan, P., Guo, L., Chen, G., Tait, A., Schneider,
  D.: Automated cattle counting using mask r-cnn in quadcopter vision system.
  Computers and Electronics in Agriculture  \textbf{171},  105300 (2020).
  \doi{https://doi.org/10.1016/j.compag.2020.105300},
  \url{http://www.sciencedirect.com/science/article/pii/S0168169919320149}

\bibitem{8599382}
{Yang}, G., {Chen}, S.: Pedestrian detection under dense crowd. In: 2018 5th
  International Conference on Systems and Informatics (ICSAI). pp. 379--382
  (2018)

\bibitem{yu19_rhnet}
Yu, R., Xu, X., Shen, Y.: Rhnet: Lightweight dilated convolutional networks for
  dense objects counting. In: 2019 Chinese Control Conference (CCC). p.~nil (7
  2019). \doi{10.23919/chicc.2019.8866393},
  \url{https://doi.org/10.23919/chicc.2019.8866393}

\bibitem{zhang2015cross}
Zhang, C., Li, H., Wang, X., Yang, X.: Cross-scene crowd counting via deep
  convolutional neural networks. In: Proceedings of the IEEE conference on
  computer vision and pattern recognition. pp. 833--841 (2015)

\bibitem{zhang16_singl_image_crowd_count_multi}
Zhang, Y., Zhou, D., Chen, S., Gao, S., Ma, Y.: Single-image crowd counting via
  multi-column convolutional neural network. In: 2016 IEEE Conference on
  Computer Vision and Pattern Recognition (CVPR) (6 2016).
  \doi{10.1109/cvpr.2016.70}, \url{https://doi.org/10.1109/cvpr.2016.70}

\end{thebibliography}

\end{document}